\documentclass[11pt]{article}
\usepackage[preprint]{acl}

\usepackage{times}
\usepackage{latexsym}
\usepackage[T1]{fontenc}
\usepackage[utf8]{inputenc}
\usepackage{microtype}
\usepackage{inconsolata}
\usepackage{graphicx}
\usepackage{booktabs}
\usepackage{multirow}
\usepackage[most]{tcolorbox}

\title{ChartDiff: A Large-Scale Benchmark for Comprehending Pairs of Charts}

\author{Rongtian Ye\\
  Department of Computer Science, Aalto University \\
  \texttt{rongtian.ye@aalto.fi}\\
  \url{https://ckchaos.github.io/ChartDiff}}

\begin{document}
\maketitle

\begin{abstract}
Charts are central to analytical reasoning, yet existing benchmarks for chart understanding focus almost exclusively on single-chart interpretation rather than comparative reasoning across multiple charts. To address this gap, we introduce \textbf{ChartDiff}, the first large-scale benchmark for cross-chart comparative summarization. ChartDiff consists of 8,541 chart pairs spanning diverse data sources, chart types, and visual styles, each annotated with LLM-generated and human-verified summaries describing differences in trends, fluctuations, and anomalies. Using ChartDiff, we evaluate general-purpose, chart-specialized, and pipeline-based models. Our results show that frontier general-purpose models achieve the highest GPT-based quality, while specialized and pipeline-based methods obtain higher ROUGE scores but lower human-aligned evaluation, revealing a clear mismatch between lexical overlap and actual summary quality. We further find that multi-series charts remain challenging across model families, whereas strong end-to-end models are relatively robust to differences in plotting libraries. Overall, our findings demonstrate that comparative chart reasoning remains a significant challenge for current vision-language models and position ChartDiff as a new benchmark for advancing research on multi-chart understanding.
\end{abstract}

\section{Introduction}
Charts play a central role in analytical reasoning, communication, and decision-making. From scientific publications to business dashboards, chart presentations allow humans to quickly extract patterns, compare trends, and evaluate alternatives. As vision-language models (VLMs)~\cite{radford2021learningtransferablevisualmodels, ee77cbab80b247598118c7c65d80fa0f, liu2023llava, Qwen-VL} continue to advance, enabling them to understand and explain charts~\cite{han2023chartllama, masry-etal-2023-unichart, zhang-etal-2024-tinychart} has become an increasingly active research area. Recent progress has produced a variety of benchmarks and methods targeting tasks such as chart question answering~\cite{masry-etal-2022-chartqa,NEURIPS2024_cdf6f8e9}, chart summarization~\cite{kantharaj-etal-2022-chart, meng-etal-2024-chartassistant}, structured information extraction~\cite{liu-etal-2023-matcha, liu-etal-2023-deplot}. Despite these advances, existing work~\cite{masry-etal-2022-chartqa,masry-etal-2023-unichart,han2023chartllama,zhang-etal-2024-tinychart, masry-etal-2025-chartqapro, meng-etal-2024-chartassistant, zhao-etal-2025-chartcoder,DBLP:conf/iclr/XuQQDXYG25} overwhelmingly focuses on single-chart understanding, treating each chart as an isolated unit.

However, many real-world analytical tasks are inherently comparative. Analysts routinely juxtapose multiple charts to evaluate differences across groups, time periods, experimental conditions, or modeling assumptions. Detecting how two charts differ—whether in their underlying data, visual encodings, statistical relationships, or narrative intent—is central to tasks such as anomaly detection, model comparison, A/B testing, monitoring system performance, or verifying the reproducibility of results. Yet, to date, though some works~\cite{zhu-etal-2025-multichartqa,iyengar-etal-2025-interchart,10.5555/3737916.3739835} have been proposed in multi-chart setting, the ability of VLMs to perform such comparative reasoning remains largely unexplored.

\begin{figure*}[t]
    \centering

    \begin{minipage}[t]{0.49\textwidth}
        \centering
        \includegraphics[width=\linewidth]{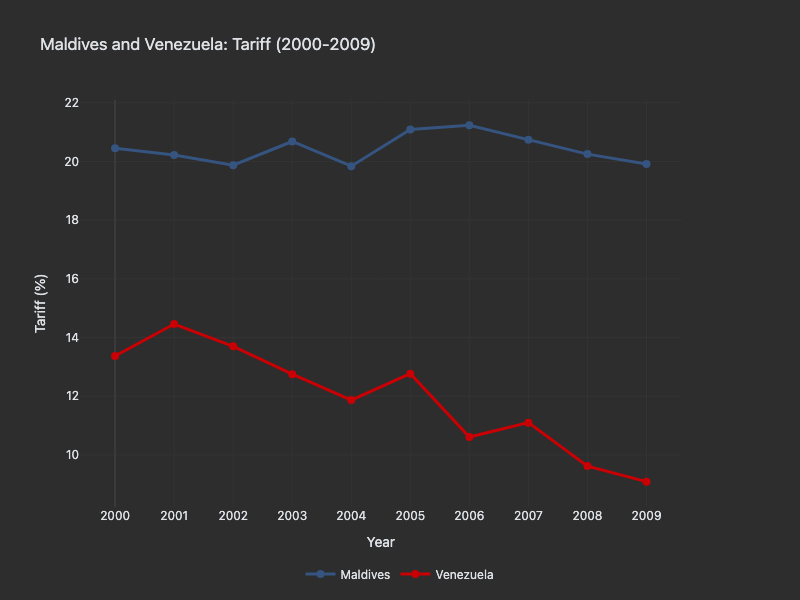}
        \small Chart A (Left)
    \end{minipage}
    \hfill
    \begin{minipage}[t]{0.49\textwidth}
        \centering
        \includegraphics[width=\linewidth]{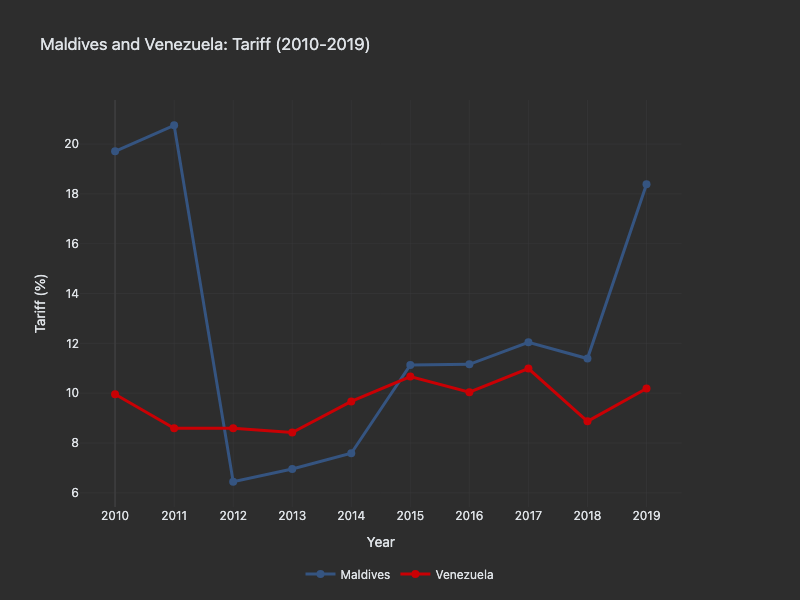}
        \small Chart B (Right)
    \end{minipage}

    \vspace{1em}

    \begin{minipage}[t]{0.98\textwidth}
        \small
        \textbf{Comparison Summary:} From 2000 to 2009, Maldives maintained a high and fairly stable tariff around 20–21\%, while Venezuela showed a clear downward trend from 13.37\% to 9.09\% with only minor fluctuations. In 2010–2019, Maldives changed much more dramatically: after staying near 20\% in 2010–2011, its tariff collapsed to 6.45\% in 2012, then gradually recovered through the decade before rebounding sharply to 18.39\% in 2019. Venezuela, by contrast, remained relatively steady in the later period, mostly fluctuating within a narrower 8.4–11.0\% range after its earlier decline. Overall, the key shift is that Maldives moved from stability to major volatility and structural change, whereas Venezuela moved from a steady decline to a comparatively stable low-tariff pattern.
    \end{minipage}
    \caption{
    \textbf{ChartDiff Dataset Illustration.}
    The task requires comparing two charts and generating a concise description of their differences. More examples can be found in Appendix~\ref{sec:dataset_examples}.
    }\label{fig:chartdiff_example}
\end{figure*}

To address this gap, we introduce a new large-scale benchmark \textbf{ChartDiff} (Dataset illustration in Figure~\ref{fig:chartdiff_example}) designed specifically for cross-chart difference comprehension. ChartDiff comprises 8,541 pairs of charts, covering a wide range of data sources, visualization types, and styles. For each pair, we provide LLM-generated and human-verified annotations describing differences in trends, fluctuations, and anomalies. These annotations form a rigorous testbed for evaluating whether VLMs can move beyond single-image interpretation and engage in comparative visual reasoning.

Using this benchmark, we conduct the first systematic evaluation of state-of-the-art VLMs on the task of cross-chart comparative summarization. Our experiments show that, while these models can achieve comparable performance on classic metrics~\cite{lin-2004-rouge}, their performance varies significantly on modern metrics~\cite{fu-etal-2024-gptscore}. Some models struggle with poor reasoning abilities. 

Our contributions are in two aspects:
\begin{itemize}
    \item We introduce the first benchmark ChartDiff on cross-chart comparative summarization, consisting of 8,541 annotated chart pairs with high diversity.
    \item We evaluate various modern VLMs on cross-chart comparative summarization to illustrate the great progress of new VLMs and provide new insights into the challenges of chart understanding.
\end{itemize}

We hope this work catalyzes future research on multi-chart reasoning, an ability that is essential for supporting real-world visual analytics workflows yet remains underrepresented in current model capabilities and benchmarks.

\section{Related Works}
\subsection{Vision-Language Models for Chart Understanding}
VLMs have made rapid progress in chart understanding, and existing approaches can be broadly divided into general-purpose multimodal models and chart-specialized models. General-purpose models such as GPT systems~\cite{openai2024gpt4ocard,singh2025openaigpt5card}, Gemini series~\cite{comanici2025gemini25pushingfrontier}, Qwen3.5~\cite{qwen3.5}, and InternVL~\cite{chen2024internvl} exhibit strong visual reasoning abilities and can handle a wide range of chart-related tasks without task-specific training. In contrast, chart-specialized models, including ChartLlama~\cite{han2023chartllama}, UniChart~\cite{masry-etal-2023-unichart}, ChartAssistant~\cite{meng-etal-2024-chartassistant}, and ChartGemma~\cite{masry-etal-2025-chartgemma} are typically trained via visual instruction tuning~\cite{liu2023llava} or multitask learning to better capture chart structures and semantics. Some works further extend encoder-decoder frameworks, such as Matcha~\cite{liu-etal-2023-matcha} built upon Pix2Struct~\cite{10.5555/3618408.3619188}, by incorporating chart data extraction and mathematical reasoning capabilities. While these models achieve strong performance on tasks like chart question answering and summarization, they are often limited by task-specific tuning and predefined pipelines, which restrict their ability to generalize to diverse chart types and complex reasoning scenarios.

Recent work focuses on improving reasoning for chart understanding. TinyChart~\cite{zhang-etal-2024-tinychart} adopts Program-of-Thought prompting~\cite{chen2022program}, while ChartCoder~\cite{zhao-etal-2025-chartcoder} and ChartReasoner~\cite{jia2025chartreasonercodedrivenmodalitybridging} leverage chart-to-code generation for multi-step inference. Chart-R1~\cite{chen2026chartr1chainofthoughtsupervisionreinforcement} and Chart-RL~\cite{zhang2026chartrlgeneralizedchartcomprehension} further incorporates reinforcement learning to enhance reasoning quality. However, these approaches depend on intermediate representations such as code or templates, making them sensitive to upstream errors and limiting robustness, as highlighted by works like ChartMimic~\cite{DBLP:conf/iclr/0002SLS0JXZLZLN25} and ChartMoE~\cite{DBLP:conf/iclr/XuQQDXYG25}.

\subsection{Benchmarks for Chart Understanding}
A wide range of benchmarks have been proposed to evaluate chart understanding capabilities of VLMs. Early datasets such as FigureQA~\cite{EbrahimiKahou2017FigureQAAA}, DVQA~\cite{kafle2018dvqa}, and PlotQA~\cite{Methani_2020_WACV} primarily rely on synthetic charts and template-based questions, which constrain both visual diversity and reasoning complexity. Subsequent efforts, including ChartQA~\cite{masry-etal-2022-chartqa} and Chart-to-Text~\cite{kantharaj-etal-2022-chart}, shift toward real-world data and more natural language queries, enabling evaluation of higher-level tasks such as question answering and summarization. More recent benchmarks, such as ChartX~\cite{Xia2024ChartXAC} and ChartBench~\cite{DBLP:journals/corr/abs-2312-15915}, further expand chart diversity and task coverage. Despite these advances, most existing benchmarks are designed around single-chart settings, where models process one visualization at a time without considering relationships across multiple charts.

To address increasing model capabilities, several works introduce more challenging tasks and broader evaluation protocols. Benchmarks like CharXiv~\cite{NEURIPS2024_cdf6f8e9} and SciGraphQA~\cite{DBLP:journals/corr/abs-2308-03349} emphasize complex reasoning, multi-turn interactions, or domain-specific knowledge, while ChartQAPro~\cite{masry-etal-2025-chartqapro} and MMMU~\cite{yue2023mmmu} explore more diverse question types and multimodal reasoning scenarios. However, these benchmarks still predominantly evaluate understanding within a single chart, focusing on tasks such as data extraction, description, or localized reasoning. As a result, they fail to capture a critical real-world requirement: comparing and synthesizing information across multiple visualizations, which often involves aligning semantics, identifying differences, and aggregating trends.

Only a limited number of benchmarks begin to explore multi-chart or cross-chart reasoning. For example, MultiChartQA~\cite{zhu-etal-2025-multichartqa} introduces multi-hop reasoning across related charts, while INTERCHART~\cite{iyengar-etal-2025-interchart} investigates cross-chart understanding under controlled settings. ReMI~\cite{10.5555/3737916.3739835} also includes a small subset of multi-chart scenarios. Nevertheless, these datasets are relatively small in scale or limited in task design, leaving substantial gaps in evaluating comprehensive cross-chart reasoning. To bridge this gap, we propose ChartDiff, a large-scale benchmark for cross-chart comparative summarization, consisting of 8,541 chart pairs. Notably, ChartAB~\cite{bansal2025chartabbenchmarkchartgrounding} presents a dataset of comparable scale, but focuses on a diagnostic framework centered on grounding, alignment, and robustness, emphasizing fine-grained pairwise difference identification rather than holistic comparative summarization. In contrast, ChartDiff explicitly targets comparative reasoning by requiring models to identify differences in trends, fluctuations, and anomalies across charts and to generate coherent summaries, offering an additional perspective for evaluating chart understanding in multi-chart scenarios.

\section{Dataset Construction}

In this section, we describe the construction pipeline of ChartDiff. We begin by collecting raw data from real-world sources, followed by preprocessing to prepare the data for chart rendering. Finally, we construct chart pairs and apply an annotation pipeline to produce comparison summaries.

\subsection{Raw Data Collection and Processing}

We collect tabular time-series data from publicly available sources, including Macrotrends~\cite{macrotrends_macrodata}, Yahoo Finance~\cite{yahoo_finance}, and Visual Crossing~\cite{visualcrossing_weather_apa}. The collected data spans eight domains: economy, health, immigration, labor force, population, trade, stock markets, and weather. In total, the dataset covers approximately 200 countries or regions, 100 cities, and 100 publicly traded stocks.

We first filter out datasets with discontinuous or incomplete time series to ensure data quality. We then perform data sampling and construct pairs of datasets for comparison. To ensure that each pair reflects a meaningful and controlled difference, we constrain paired datasets to differ along only one of the following three aspects: (1) data entity (e.g., different countries or stocks), (2) time span, or (3) data category.

After pairing, each pair consists of two CSV datasets, which are used to generate a pair of charts for subsequent comparison and annotation.

\subsection{Chart Rendering}

To generate visualizations, we utilize three widely used Python plotting libraries: Matplotlib~\cite{Hunter:2007}, Plotly~\cite{plotly}, and Plotnine~\cite{plotnine_github}. For each library, we design multiple styling configurations to enhance visual diversity, resulting in approximately 60 distinct visualization styles in total.

ChartDiff includes six chart categories: line charts, bar charts, horizontal bar charts, multi-series line charts\footnote{A multi-series chart is defined as a visualization where the underlying CSV data contains multiple distinct value columns. An example pair of multi-series line charts is shown in Figure~\ref{appendix_fig:chartdiff_example_linem}, and an example pair of multi-series bar charts is shown in Figure~\ref{appendix_fig:chartdiff_example_barm}.}, multi-series bar charts, and pie charts. These chart types cover a wide range of common visualization scenarios, including temporal trends, categorical comparisons, and proportional distributions.

To ensure high-quality visualizations, all generated charts undergo manual inspection. We verify the absence of common visualization issues, including legend occlusion, missing or improperly rendered data points, inconsistent axis scaling, and other artifacts that could hinder accurate interpretation.

\subsection{Annotation Pipeline}
To produce high-quality comparison summaries, we employ a multi-stage annotation pipeline leveraging large language models (LLMs). The pipeline follows an \textit{annotate--judge--verify} paradigm for each chart pair.

First, we define a pool of LLMs $\mathcal{A}$. For each chart pair $P$, we randomly sample an annotator model $L_1 \in \mathcal{A}$ and prompt it with a carefully designed instruction(Figure~\ref{prompt_for_candidate_annotations}) to generate a candidate comparison summary $S$. We provide only the underlying CSV data of the charts to ensure accurate and consistent analysis.

Next, we sample a second model $L_2 \in \mathcal{A} \setminus \{L_1\}$ to act as a judge. Given the same chart pair $P$ and the candidate summary $S$, $L_2$ evaluates whether $S$ is accurate and acceptable according to a predefined evaluation prompt(Figure~\ref{prompt_for_review_annotations}), and decides whether to accept or reject it. For accepted summaries, the words “Dataset” and “dataset” are replaced with “Chart” and “chart”, respectively.

Finally, all accepted summaries undergo manual verification to ensure quality. During this stage, we check for factual correctness, completeness of key differences, and overall clarity, filtering out any remaining low-quality or ambiguous annotations.

\subsection{Dataset Curation Details}

For data pairing, we sample between 6 and 12 data points for each dataset across all chart types, except for pie charts, and ensure that the two datasets in each pair contain the same number of data points. For pie charts, we restrict each dataset to 3--5 categories to maintain clear and interpretable proportional comparisons.

For chart rendering, we randomly select a visualization style from a set of predefined styling configurations for each pair, and both charts in the pair share the same styling configuration.

In the annotation pipeline, we employ a pool of LLMs, including GPT-5.4~\cite{singh2025openaigpt5card} and Gemini 3.1 Pro~\cite{gemini_3.1_pro}. The acceptance rate of candidate summaries generated by GPT-5.4 is 0.93, while that of Gemini 3.1 Pro is 0.967, indicating a high level of annotation quality.

After completing the three-stage pipeline, we obtain a total of 8,541 chart pairs with high-quality comparison annotations. The distribution of samples across different chart types is summarized in Table~\ref{tab:plot_type_stats}.

Each chart is rendered at a resolution of $800 \times 600$ pixels. We further split ChartDiff into training, validation, and test sets containing 6,041, 1,000, and 1,500 chart pairs, respectively.

\begin{table}
  \centering
  \begin{tabular}{lcc}
    \toprule
    \textbf{Chart Type} & \textbf{Pairs} &  \textbf{Percentage}\\
    \hline
    Line    & 2257  &    26.4\%     \\
    Bar    & 2153   &     25.2\%   \\
    Horizontal Bar    & 679  &  8.0\%        \\
    Line (Multi-series) & 1044 &12.2\%\\
    Bar (Multi-series) & 1072 &12.6\%\\
    Pie & 1336 &15.6\%\\
    \bottomrule
  \end{tabular}
  \caption{Distribution of chart pairs across different chart types in ChartDiff.}
  \label{tab:plot_type_stats}
\end{table}

\section{Experiments}

\subsection{Models}

We evaluate a diverse set of models spanning four categories:

\textbf{General-Purpose Closed-Source Models.}
We include state-of-the-art proprietary VLMs, including GPT-5.4~\cite{singh2025openaigpt5card}, Gemini 3.1 Pro~\cite{gemini_3.1_pro}, GPT-5.4-mini~\cite{singh2025openaigpt5card}, Gemini 3.1 Flash Lite~\cite{gemini_3.1_flash_Lite}, Claude Sonnet 4.6~\cite{claude_sonnet_4.6}, and GPT-4o~\cite{openai2024gpt4ocard}.

\textbf{General-Purpose Open-Source Models.}
We consider representative open-source models, including Qwen3.5-397B-A17B~\cite{qwen3.5}, Qwen3.5-9B~\cite{qwen3.5}, and Qwen2.5-VL-7B~\cite{Bai2025Qwen25VLTR}.

\textbf{Chart-Domain Specialized Models.}
We evaluate models specifically designed for chart understanding, including ChartGemma~\cite{masry-etal-2025-chartgemma} and MatCha~\cite{liu-etal-2023-matcha}.

\textbf{Pipeline-Based Methods.}
In addition to end-to-end models, we evaluate a pipeline-based approach that first extracts structured data from charts and then performs comparison using an LLM. Specifically, we use DePlot~\cite{liu-etal-2023-deplot} as the chart-to-table extractor, followed by GPT-5.4~\cite{singh2025openaigpt5card} or Qwen3.5-9B~\cite{qwen3.5} for comparison generation.

Finally, to establish a lower-bound baseline, we prompt GPT-5.4~\cite{singh2025openaigpt5card} to randomly generated outputs (Figure~\ref{prompt_random_output}), which serve as a reference for non-informative predictions.

\subsection{Evaluation Metrics}

We adopt two complementary evaluation metrics:

\textbf{ROUGE.}
We use ROUGE~\cite{lin-2004-rouge} as a standard lexical-overlap metric to measure similarity between generated summaries and reference annotations.

\textbf{GPT Score.}
We further employ a model-based evaluation metric, GPT Score~\cite{fu-etal-2024-gptscore}, using GPT-5.4~\cite{singh2025openaigpt5card} as the judge model to assess the quality of generated summaries with a predefined grading prompt (Figure~\ref{prompt_gpt_score_part_1} and Figure~\ref{prompt_gpt_score_part_2}).

To validate the reliability of GPT Score, we randomly sample 300 comparison summaries generated by different models and obtain human ratings using the same evaluation criteria as the grading prompt (Figure~\ref{prompt_gpt_score_part_1} and Figure~\ref{prompt_gpt_score_part_2}). We observe a Pearson correlation coefficient of 0.91 between human scores and GPT scores, indicating strong agreement.

\begin{table*}
  \centering
  \small
  \begin{tabular}{lcccc}
    \toprule
    \textbf{Models}  &
    \textbf{ROUGE-1} & \textbf{ROUGE-2} & \textbf{ROUGE-L} & \textbf{GPT Score}\\
    \toprule
    \multicolumn{5}{c}{\textit{General-Purpose Closed-Source Models} }\\
    \hline
    GPT-5.4 \cite{singh2025openaigpt5card} &46.02& 12.28& 23.45& \textbf{4.95}\\
    Gemini 3.1 Pro \cite{gemini_3.1_pro} & 47.21& \textbf{13.48}& \textbf{24.2}& 4.86\\
    GPT-5.4-mini \cite{singh2025openaigpt5card}  & 43.0& 10.62& 21.68& 4.82\\
    Gemini 3.1 Flash Lite\cite{gemini_3.1_flash_Lite} & 46.37& 12.83& 22.82& 4.63\\
    Claude Sonnet 4.6 \cite{claude_sonnet_4.6} & \textbf{47.54}& 13.31& 23.42& 4.58\\
    GPT-4o \cite{openai2024gpt4ocard} & 44.43& 11.48& 22.44& 4.23\\
    \hline
    \multicolumn{5}{c}{\textit{General-Purpose Open-Source Models} }\\
    \hline
    Qwen3.5-397B-A17B \cite{qwen3.5} & \textbf{47.07}& \textbf{12.68}& \textbf{22.57}& \textbf{4.54} \\
    Qwen3.5-9B \cite{qwen3.5} & 44.09& 10.84& 21.16& 3.65\\
    Qwen2.5VL-7B \cite{Bai2025Qwen25VLTR} & 41.18& 9.82& 20.88& 3.18\\
    
    \hline
    \multicolumn{5}{c}{\textit{Chart-Domain Specialized Models} }\\
    \hline
    ChartGemma \cite{masry-etal-2025-chartgemma}  & \textbf{51.49}& 17.81& 28.53& \textbf{2.0}\\
    MatCha \cite{liu-etal-2023-matcha} & 49.52& \textbf{18.34}& \textbf{28.75}& 1.45\\
    \hline
    \multicolumn{5}{c}{\textit{Pipeline-Based Methods} }\\
    \hline
    Deplot \cite{liu-etal-2023-deplot} + GPT-5.4  & \textbf{50.75}& \textbf{17.25}& \textbf{28.88}& \textbf{3.58}\\
    Deplot + GPT-4o  & 46.46& 13.19& 23.66& 3.38\\
    Deplot + Qwen3.5-9B  & 43.1& 10.38& 20.3& 2.81\\
    \hline
    Random &  25.5& 2.5& 12.81& 1.17\\
    \bottomrule
  \end{tabular}
  \caption{
    Performance comparison of all evaluated models. Boldface indicates the best result within each group.
  }\label{model_comparison}
\end{table*}

\begin{table*}
  \centering
  \small
  \begin{tabular}{lccccccc}
    \toprule
    \multirow{3}{*}{\textbf{Models} }  &  \multicolumn{7}{c}{\textbf{GPT Score}}\\
    \cmidrule(lr){2-8}
     & \multirow{2}{*}{\textbf{Overall}} & \multicolumn{6}{c}{\textbf{Chart Type}} \\
    \cmidrule(lr){3-8} 
    & & \textbf{Line}  & \textbf{Bar}           & \textbf{Bar(H.)} & \textbf{Line(M.)} & \textbf{Bar(M.)} & \textbf{Pie} \\
    \toprule
    \multicolumn{8}{c}{\textit{ General-Purpose Closed-Source Models} }\\
    \hline
    GPT-5.4 \cite{singh2025openaigpt5card} & \textbf{4.95}& \textbf{4.97}& \textbf{4.97}& 4.89& \textbf{4.9}& \textbf{4.88}& \textbf{4.99}\\
    Gemini 3.1 Pro \cite{gemini_3.1_pro} & 4.86& 4.82& 4.9& \textbf{4.94}& 4.65& 4.85& 4.98\\
    GPT-5.4-mini \cite{singh2025openaigpt5card} & 4.82& 4.86& 4.82& 4.75& 4.74& 4.78& 4.85\\
    Gemini 3.1 Flash Lite \cite{gemini_3.1_flash_Lite}  & 4.63& 4.65& 4.66& 4.67& 4.33& 4.47& 4.87\\
    Claude Sonnet 4.6 \cite{claude_sonnet_4.6} & 4.58& 4.54& 4.6& 4.57& 4.38& 4.46& 4.87\\
    GPT-4o \cite{openai2024gpt4ocard} & 4.23& 4.23& 4.32& 3.85& 3.88& 3.78& 4.85\\
    \hline
    \multicolumn{8}{c}{\textit{ General-Purpose Open-Source Models} }\\
    \hline
    Qwen3.5-397B-A17B \cite{qwen3.5} & \textbf{4.54}& \textbf{4.59}& \textbf{4.58}& \textbf{4.36}& \textbf{4.41}& \textbf{4.35}& \textbf{4.72}\\
    Qwen3.5-9B \cite{qwen3.5} & 3.65& 3.82& 3.89& 3.55& 3.2& 3.33& 3.57\\
    Qwen2.5VL-7B \cite{Bai2025Qwen25VLTR} & 3.18& 3.54& 3.14& 2.79& 2.79& 2.53& 3.58\\
    \hline
    \multicolumn{8}{c}{\textit{Chart-Domain Specialized Models} }\\
    \hline
    ChartGemma \cite{masry-etal-2025-chartgemma} & \textbf{2.0}& \textbf{2.36}& \textbf{2.36}& \textbf{2.01}& \textbf{1.3}& \textbf{1.36}& \textbf{1.68}\\
    MatCha \cite{liu-etal-2023-matcha}& 1.45& 1.62& 1.69& 1.43& 0.92& 0.99& 1.47\\
    \hline
    \multicolumn{8}{c}{\textit{Pipeline-Based Methods} }\\
    \hline
    Deplot \cite{liu-etal-2023-deplot} + GPT-5.4& \textbf{3.58}& \textbf{3.89}& \textbf{4.63}& 3.16& \textbf{2.91}& \textbf{4.65}& 1.24\\
    Deplot + GPT-4o & 3.38& 3.68& 4.32& \textbf{3.23}& 2.77& 4.21& \textbf{1.27}\\
    Deplot + Qwen3.5-9B & 2.81& 3.16& 3.8& 2.16& 2.3& 3.44& 0.79\\
    \hline
    Random  & 1.17& 1.23& 1.5& 1.52& 1.28& 1.24& 0.21\\
    \bottomrule
  \end{tabular}
  \caption{
    GPT Score results by chart type for all evaluated models. Bar(H.) indicates horizontal bar charts. Line(M.) indicates multi-series line charts. Bar(M.) indicates multi-series bar charts. Boldface indicates the best result within each model category.
  }\label{plot_type_result}
\end{table*}

\begin{table*}
  \centering
  \begin{tabular}{lcccc}
    \toprule
    \multirow{2}{*}{\textbf{Models} }   & \multirow{2}{*}{\textbf{Overall}} & \multicolumn{3}{c}{\textbf{Plotting Libraries}} \\
    \cmidrule(lr){3-5} 
    & & \textbf{Matplotlib}  & \textbf{Plotly}           & \textbf{Plotnine} \\
    \hline
    GPT-5.4 \cite{singh2025openaigpt5card}& 4.95& 4.94& 4.97& 4.93\\
    Qwen3.5-397B-A17B \cite{qwen3.5}& 4.54& 4.51&4.63& 4.44\\
    ChartGemma \cite{masry-etal-2025-chartgemma} & 2.0& 2.06&1.99&1.95\\
    Deplot \cite{liu-etal-2023-deplot}+ GPT-5.4& 3.58& 4.08& 3.12 & 3.89\\

    \bottomrule
  \end{tabular}
  \caption{
GPT Score of representative models across different plotting libraries.
  }\label{plot_lib_result}
\end{table*}

\subsection{Implementation Details}
For all models, we concatenate the two charts horizontally into a single combined image as the model input.

For general-purpose models, we evaluate performance in a zero-shot setting without task-specific fine-tuning. We directly prompt the models (Figure~\ref{prompt_for_genrate_predictions}) to generate comparison summaries.

For pipeline-based methods, we first use a pretrained DePlot~\cite{liu-etal-2023-deplot} model for chart-to-table extraction without additional fine-tuning, and then prompt the selected LLM (Figure~\ref{prompt_for_genrate_predictions_in_pipeline}) with the extracted tables to generate comparison summaries.

For chart-specific models, including ChartGemma~\cite{masry-etal-2025-chartgemma} and MatCha~\cite{liu-etal-2023-matcha}, we fine-tune each model on the ChartDiff training set for five epochs.

\subsection{Analysis}
\textbf{Overall.} Table~\ref{model_comparison} shows that general-purpose closed-source models achieve the best overall generation quality, with GPT-5.4~\cite{singh2025openaigpt5card} obtaining the highest GPT Score (4.95), followed by Gemini 3.1 Pro~\cite{gemini_3.1_pro} (4.86). In contrast, chart-domain specialized models and pipeline-based methods achieve the strongest ROUGE scores, with ChartGemma~\cite{masry-etal-2025-chartgemma} reaching the highest ROUGE-1 (51.49), MatCha~\cite{liu-etal-2023-matcha} the best ROUGE-2 (18.34) and ROUGE-L (28.75), and Deplot~\cite{liu-etal-2023-deplot} + GPT-5.4 also performing strongly on all ROUGE metrics. However, these models obtain much lower GPT Scores, indicating a substantial mismatch between lexical overlap and human-aligned generation quality. Among open-source models, Qwen3.5-397B-A17B~\cite{qwen3.5} is the strongest, achieving competitive ROUGE scores and a GPT Score of 4.54, though it still trails the best proprietary systems. Overall, the results suggest that while specialized and pipeline-based approaches are advantageous for reference matching, frontier general-purpose LLMs produce more natural and better-evaluated chart descriptions, highlighting the importance of complementing lexical overlap-based metrics with quality-oriented evaluation metrics.

\textbf{Chart type.} Table~\ref{plot_type_result} shows clear performance differences across chart types. Pie charts are generally the easiest for general-purpose LLMs, with nearly all closed-source models achieving very high GPT Scores on this type. Simple line and bar charts are also handled well, while multi-series charts are more challenging, as performance tends to drop across most model families. This pattern is particularly evident for smaller open-source and chart-specialized models, whose scores decrease substantially on multi-series charts. In contrast, pipeline-based methods perform relatively well on bar and multi-series bar charts but fail badly on pie charts, suggesting that their effectiveness is highly dependent on chart structure, possibly because DePlot was not pretrained on pie charts. Overall, the results indicate that chart complexity is a major factor in chart-to-text generation, with multi-series charts remaining the most difficult cases.

\textbf{Plotting library.} Table~\ref{plot_lib_result} shows that performance is generally stable across plotting libraries for strong end-to-end models, suggesting that library-specific rendering differences have limited impact on top-performing LLMs. For example, GPT-5.4~\cite{singh2025openaigpt5card} maintains nearly identical GPT Scores on Matplotlib (4.94), Plotly (4.97), and Plotnine (4.93), while Qwen3.5-397B-A17B also shows only moderate variation across libraries. Among the three libraries, Plotly appears slightly easier for general-purpose models, yielding the highest scores for both GPT-5.4 and Qwen3.5-397B-A17B~\cite{qwen3.5}. In contrast, pipeline-based methods are more sensitive to library choice: Deplot~\cite{liu-etal-2023-deplot} + GPT-5.4 performs best on Matplotlib (4.08) and Plotnine (3.89) but drops noticeably on Plotly (3.12). Overall, the results suggest that plotting library has only a minor effect for strong end-to-end LLMs, but can substantially affect modular pipeline approaches.

\section{Conclusion}
\label{sec:conclusion}
We present ChartDiff, the first large-scale benchmark for cross-chart comparative summarization, consisting of 8,541 chart pairs with high-quality comparison annotations. Our evaluation shows that although modern vision-language models have become strong at single-chart understanding, they still face clear challenges in comparative chart reasoning, especially on structurally complex chart types such as multi-series charts. We further find a substantial mismatch between lexical-overlap metrics and GPT-based quality evaluation, suggesting that chart comparison should be assessed with more than overlap-based metrics alone. We hope ChartDiff will provide a useful foundation for future research on multi-chart understanding and comparative visual reasoning.

\section*{Limitations}
Our work has several limitations. ChartDiff covers only a subset of common chart types and may not generalize to more complex real-world visualizations. Although annotations are human-verified, they are partially LLM-generated and may reflect annotation bias. In addition, our main evaluation relies on GPT-Score, which, despite strong correlation with human judgment, is still an imperfect automatic evaluator. Finally, we focus on open-ended comparative summarization rather than other multi-chart reasoning tasks, leaving broader comparative understanding for future work.

\section*{Ethics Statement}
ChartDiff is constructed from publicly available data and synthetically generated charts and does not contain personal or sensitive information. The dataset is intended for research purposes only. We note that models evaluated on this benchmark may generate incorrect or misleading summaries, and therefore should not be used in high-stakes applications without human verification. We encourage responsible use of the dataset and future work on improving the reliability and safety of chart understanding systems.

\section*{Acknowledgements}
The author thanks the anonymous reviewers for their careful reading and constructive feedback.

The author is deeply grateful to his friends FY, Dingyuan, and Xinchi for their unwavering support throughout this research. Their constant encouragement, together with their equally constant insults, proved invaluable in sustaining this work.

The author also acknowledges the computational resources provided by CSC -- IT Center for Science, Finland.
\bibliography{custom}

\clearpage
\appendix
\onecolumn
\label{sec:appendix}
\section{Dataset Examples}
\label{sec:dataset_examples}

\begin{figure*}[h]
  \centering
  \includegraphics[width=\linewidth]{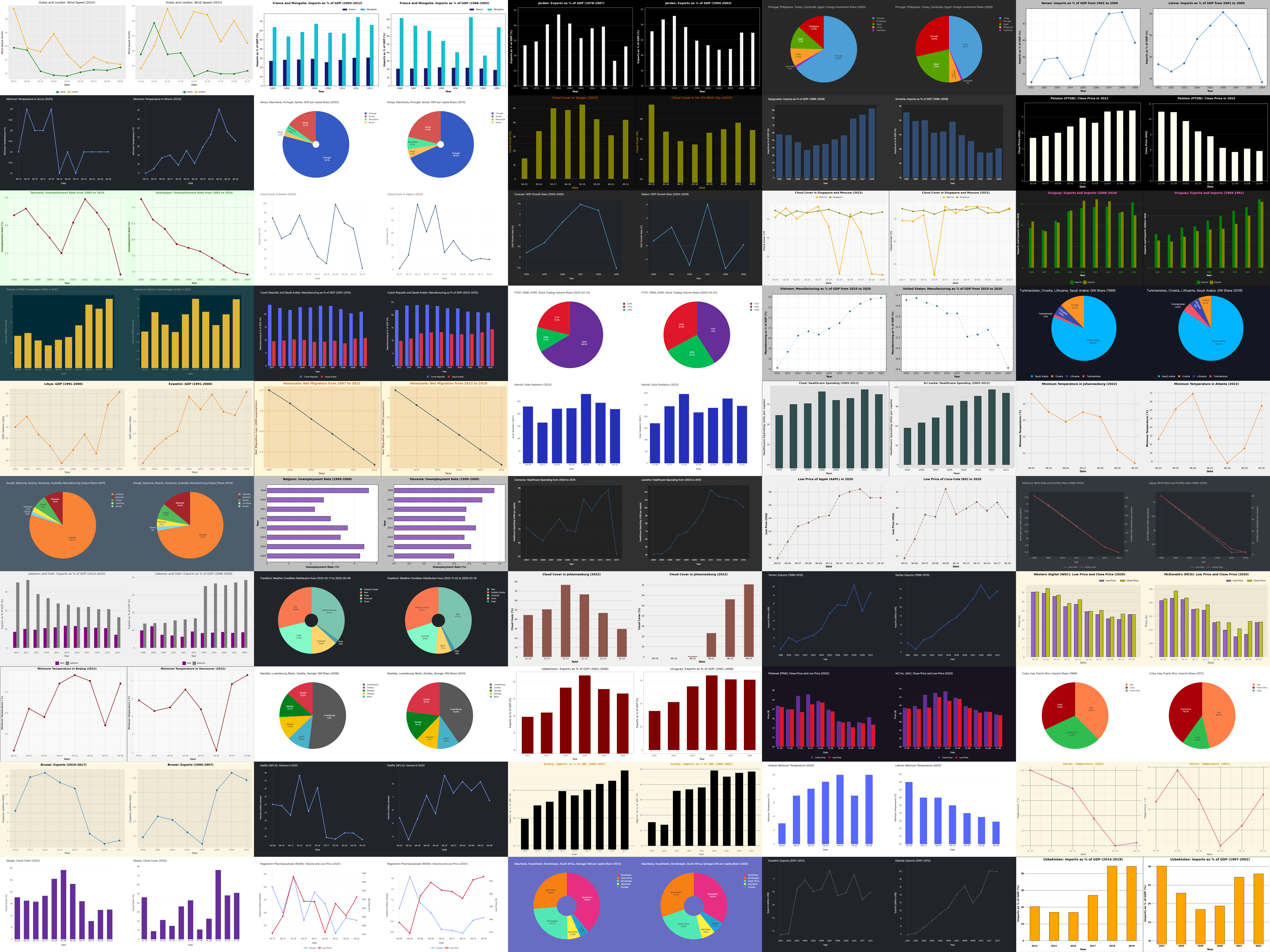}
  \caption{Fifty randomly selected chart pairs from the ChartDiff dataset.}
\end{figure*}

\begin{figure*}
    \centering

    \begin{minipage}[t]{0.49\textwidth}
        \centering
        \includegraphics[width=\linewidth]{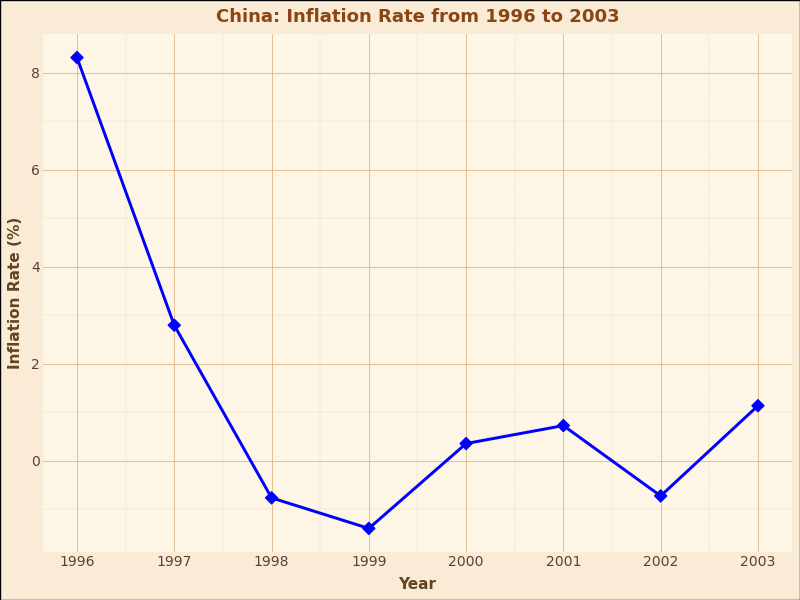}
        \small Chart A (Left)
    \end{minipage}
    \hfill
    \begin{minipage}[t]{0.49\textwidth}
        \centering
        \includegraphics[width=\linewidth]{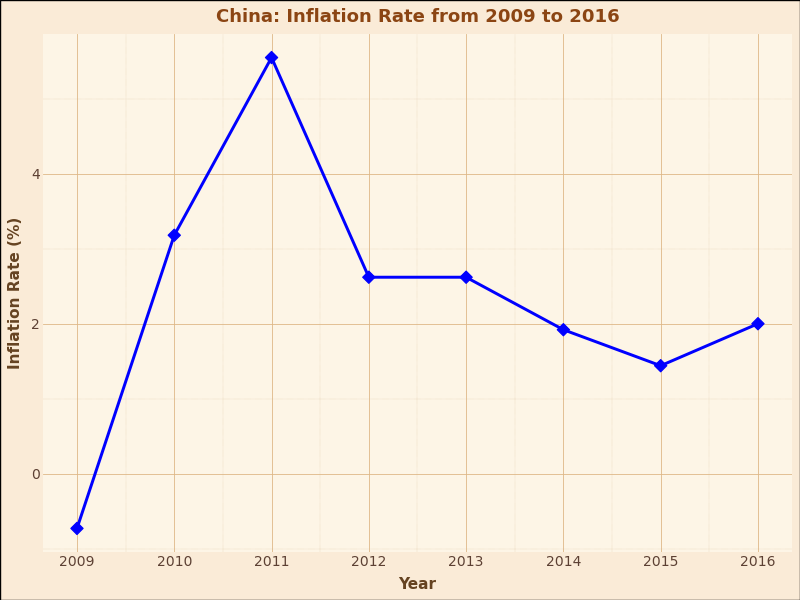}
        \small Chart B (Right)
    \end{minipage}

    \vspace{1em}

    \begin{minipage}[t]{0.98\textwidth}
        \small
        \textbf{Comparison Summary:} A comparison of China's inflation rates between 1996–2003 and 2009–2016 reveals distinct differences in economic volatility and overall trends. The 1996–2003 period was characterized by a dramatic plunge from an initial high of 8.31\% to recurring deflationary episodes, hitting a low of -1.4\% in 1999 and remaining near or below zero for most of the timeframe. In contrast, the 2009–2016 period began with slight deflation at -0.73\% but quickly spiked to a peak of 5.55\% in 2011 before gradually cooling and stabilizing around 2\%. Ultimately, while the earlier dataset highlights a persistent downward trajectory into multiple years of negative inflation, the later dataset demonstrates a rapid post-deflation recovery followed by sustained, moderate positive inflation.
    \end{minipage}
    \caption{
    An example pair of line charts.
    }\label{appendix_fig:chartdiff_example_line}
\end{figure*}

\begin{figure*}[t]
    \centering

    \begin{minipage}[t]{0.49\textwidth}
        \centering
        \includegraphics[width=\linewidth]{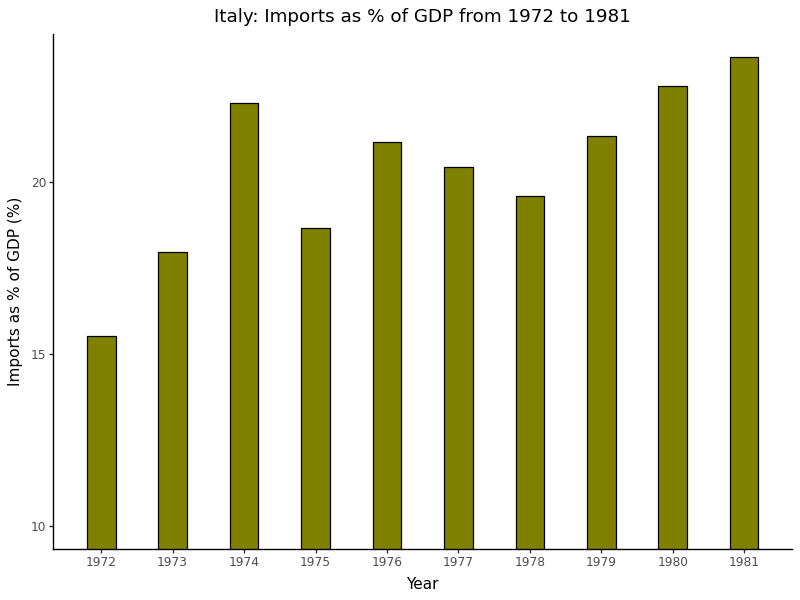}
        \small Chart A (Left)
    \end{minipage}
    \hfill
    \begin{minipage}[t]{0.49\textwidth}
        \centering
        \includegraphics[width=\linewidth]{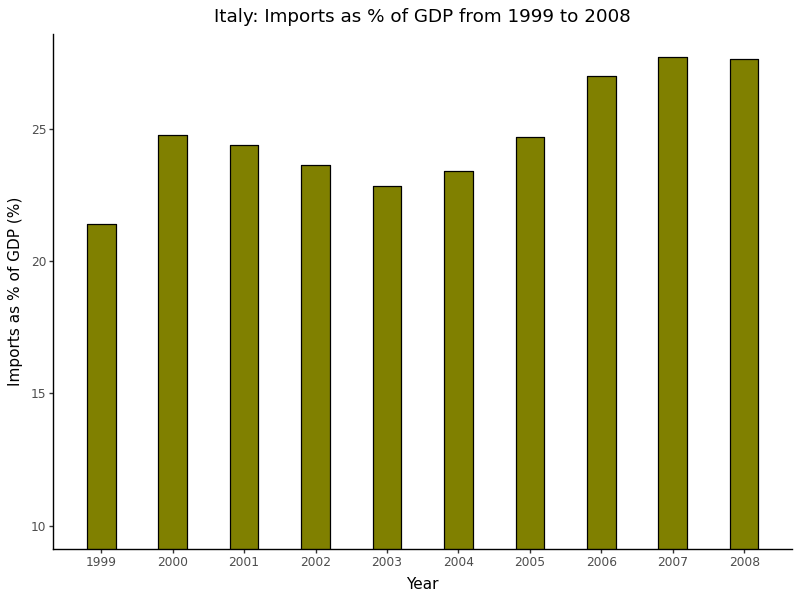}
        \small Chart B (Right)
    \end{minipage}

    \vspace{1em}

    \begin{minipage}[t]{0.98\textwidth}
        \small
        \textbf{Comparison Summary:} A comparison of Italy's imports as a percentage of GDP between the 1972–1981 and 1999–2008 periods reveals a substantially higher baseline for imports in the later decade. During the 1970s, the import share started at a low of 15.51\% and experienced significant volatility, notably spiking to 22.28\% in 1974 before dropping sharply to 18.67\% the following year. Conversely, the 1999–2008 period operated at an elevated level, beginning at 21.42\% and demonstrating a much smoother overall growth trajectory. This later decade saw steady annual increases from 2003 onward, peaking at 27.70\% in 2007 before slightly plateauing in 2008. Despite the differences in volatility and baseline values, both datasets ultimately share a consistent long-term upward trend in Italy's import-to-GDP ratio over their respective ten-year spans.
    \end{minipage}
    \caption{
    An example pair of bar charts.
    }\label{appendix_fig:chartdiff_example_bar}
\end{figure*}

\begin{figure*}[t]
    \centering

    \begin{minipage}[t]{0.49\textwidth}
        \centering
        \includegraphics[width=\linewidth]{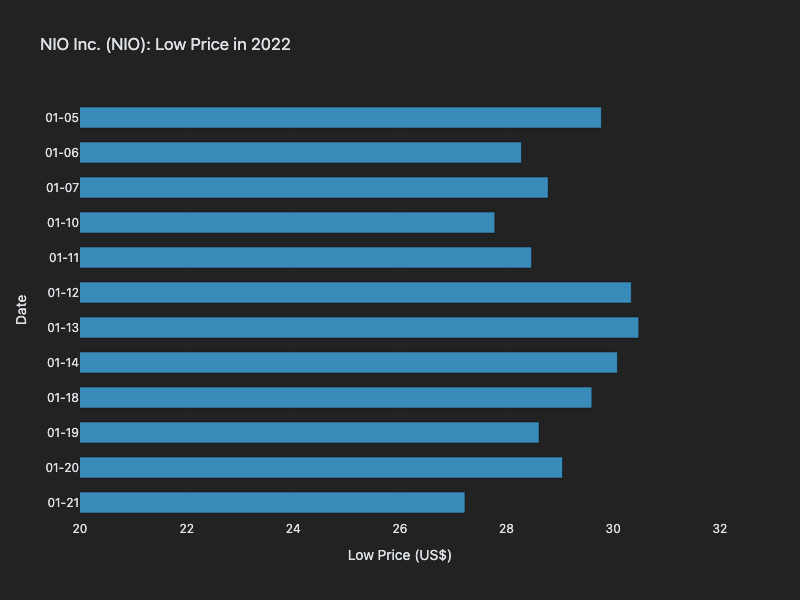}
        \small Chart A (Left)
    \end{minipage}
    \hfill
    \begin{minipage}[t]{0.49\textwidth}
        \centering
        \includegraphics[width=\linewidth]{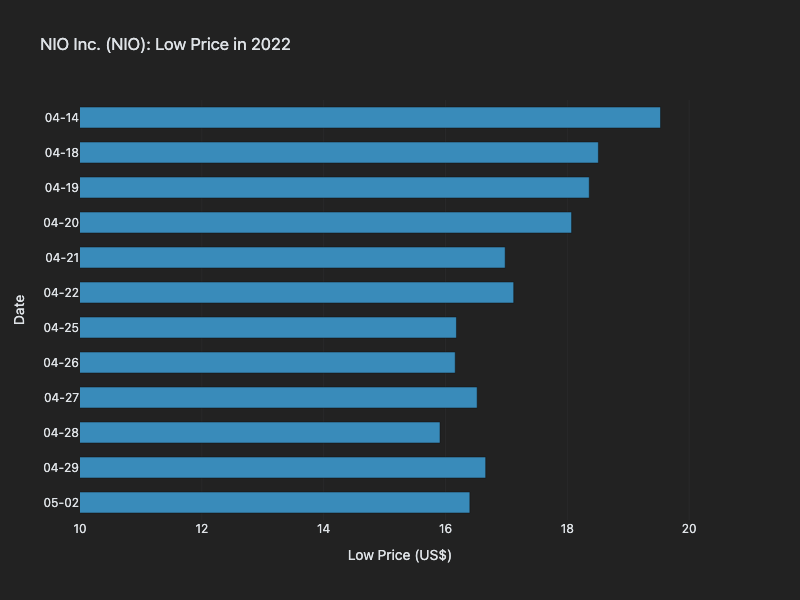}
        \small Chart B (Right)
    \end{minipage}

    \vspace{1em}

    \begin{minipage}[t]{0.98\textwidth}
        \small
        \textbf{Comparison Summary:} A comparison of NIO Inc.'s daily low stock prices in 2022 reveals a significant depreciation in the stock's value between January and April. In January, the low prices fluctuated at a relatively high baseline, ranging from \$27.22 to a peak of \$30.48 mid-month. By contrast, the late April to early May period shows that prices had plummeted to a much lower bracket of \$15.91 to \$19.53. Furthermore, while the January data displayed mild volatility centered around the \$28 to \$30 mark, the April data experienced a consistent downward trajectory that bottomed out at \$15.91 on April 28. Overall, the datasets highlight a drastic downward shift in NIO's market valuation over the first four months of the year.
    \end{minipage}
    \caption{
    An example pair of horizontal bar charts.
    }\label{appendix_fig:chartdiff_example_bh}
\end{figure*}

\begin{figure*}[t]
    \centering

    \begin{minipage}[t]{0.49\textwidth}
        \centering
        \includegraphics[width=\linewidth]{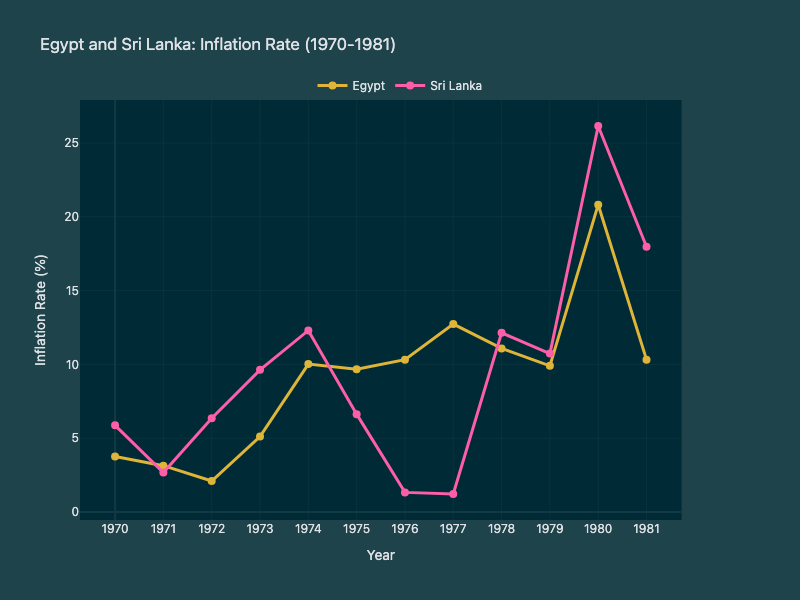}
        \small Chart A (Left)
    \end{minipage}
    \hfill
    \begin{minipage}[t]{0.49\textwidth}
        \centering
        \includegraphics[width=\linewidth]{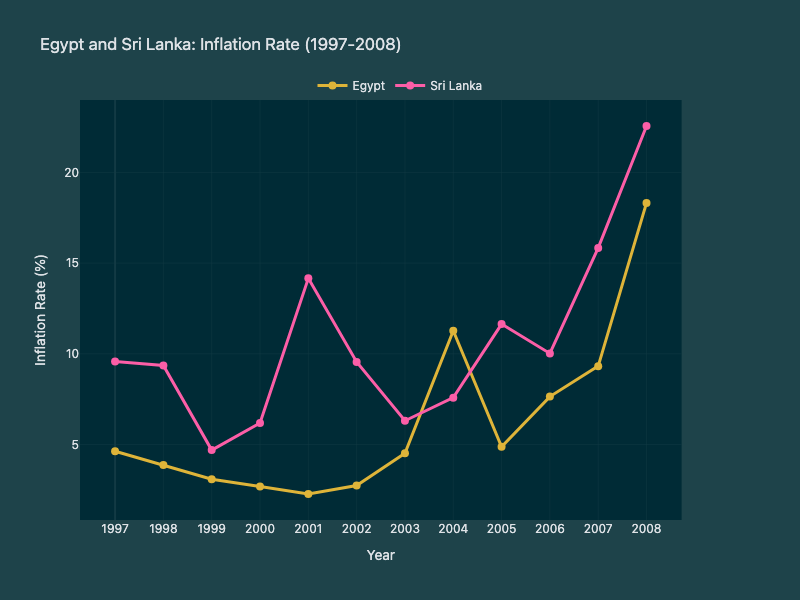}
        \small Chart B (Right)
    \end{minipage}

    \vspace{1em}

    \begin{minipage}[t]{0.98\textwidth}
        \small
        \textbf{Comparison Summary:} Across both periods, Egypt and Sri Lanka show generally moderate inflation punctuated by occasional spikes, but the earlier period is more volatile overall, especially for Sri Lanka. From 1970–1981, Egypt rises from low single digits to a late surge in 1980 before easing in 1981, while Sri Lanka swings much more sharply, dropping to near 1\% in 1976–1977 before jumping above 12\% in 1978 and peaking at 26.15\% in 1980. In 1997–2008, Egypt is comparatively stable at low inflation through the late 1990s and early 2000s, then trends upward after 2003 to 18.32\% in 2008, whereas Sri Lanka remains consistently higher than Egypt in most years and shows repeated double-digit readings, notably in 2001, 2005, 2007, and 2008. A key difference is that the first period features deeper troughs and more abrupt reversals, especially for Sri Lanka, while the second period shows a clearer upward inflation trend for both countries toward 2008. Overall, Sri Lanka tends to have higher and more erratic inflation than Egypt in both datasets, with both countries ending each period with elevated inflation.
    \end{minipage}
    \caption{
    An example pair of multi-series line charts.
    }\label{appendix_fig:chartdiff_example_linem}
\end{figure*}

\begin{figure*}[t]
    \centering

    \begin{minipage}[t]{0.49\textwidth}
        \centering
        \includegraphics[width=\linewidth]{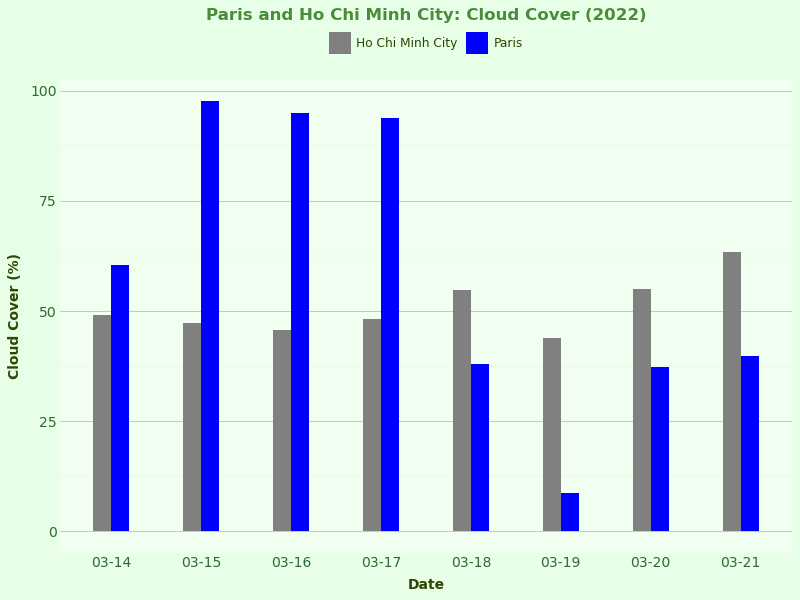}
        \small Chart A (Left)
    \end{minipage}
    \hfill
    \begin{minipage}[t]{0.49\textwidth}
        \centering
        \includegraphics[width=\linewidth]{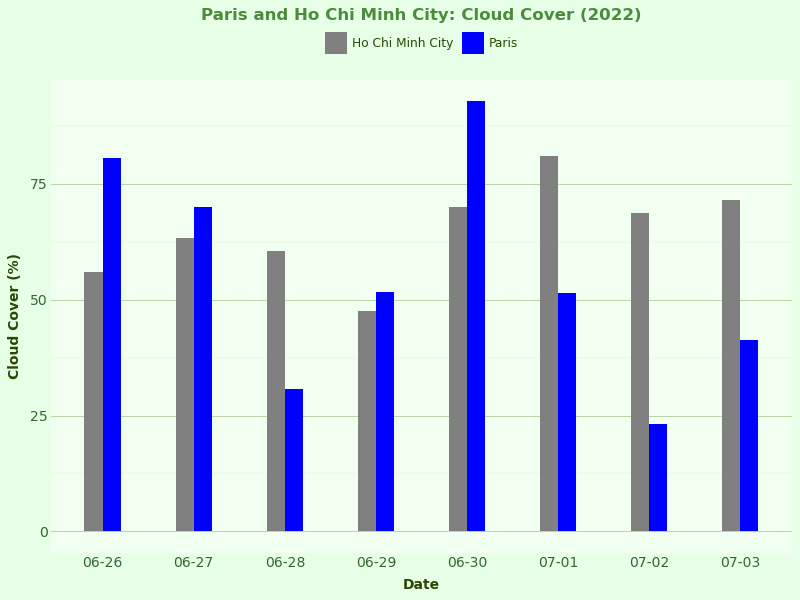}
        \small Chart B (Right)
    \end{minipage}

    \vspace{1em}

    \begin{minipage}[t]{0.98\textwidth}
        \small
        \textbf{Comparison Summary:} A comparison of cloud cover between Paris and Ho Chi Minh City across two periods in 2022 reveals distinct volatility patterns and seasonal shifts. In mid-March, Paris experienced extreme fluctuations, with cloud cover plummeting from a near-overcast peak of 97.6\% to a low of 8.8\%, while Ho Chi Minh City remained relatively stable between 43.9\% and 63.4\%. Conversely, during the late June to early July period, Ho Chi Minh City experienced an overall increase in cloudiness, trending upward to a peak of 81.0\% on July 1. Paris continued to show high atmospheric volatility during this summer timeframe, oscillating wildly between a low of 23.2\% and a high of 92.8\%. Overall, while Paris consistently demonstrated erratic, sharp shifts in both timeframes, Ho Chi Minh City transitioned from moderate, consistent cloud cover in the spring to much cloudier conditions in the summer.
    \end{minipage}
    \caption{
    An example pair of multi-series bar charts.
    }\label{appendix_fig:chartdiff_example_barm}
\end{figure*}

\begin{figure*}[t]
    \centering

    \begin{minipage}[t]{0.49\textwidth}
        \centering
        \includegraphics[width=\linewidth]{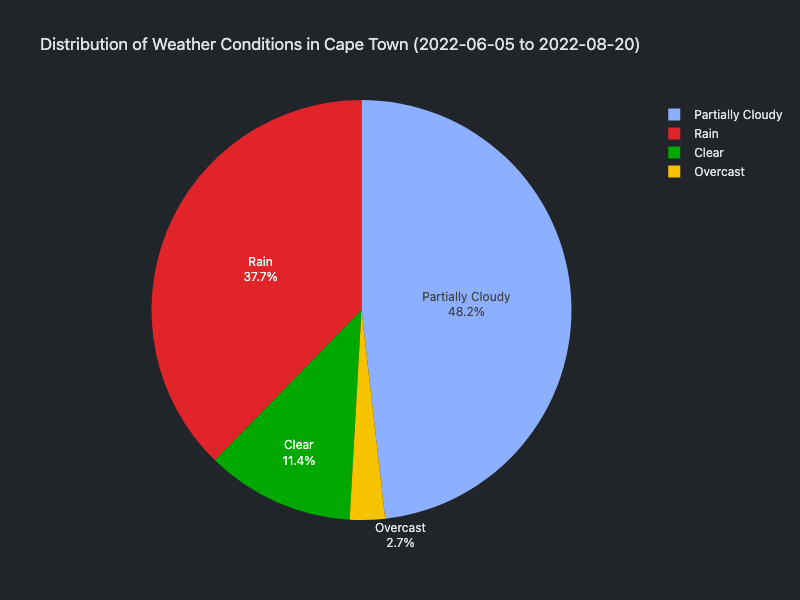}
        \small Chart A (Left)
    \end{minipage}
    \hfill
    \begin{minipage}[t]{0.49\textwidth}
        \centering
        \includegraphics[width=\linewidth]{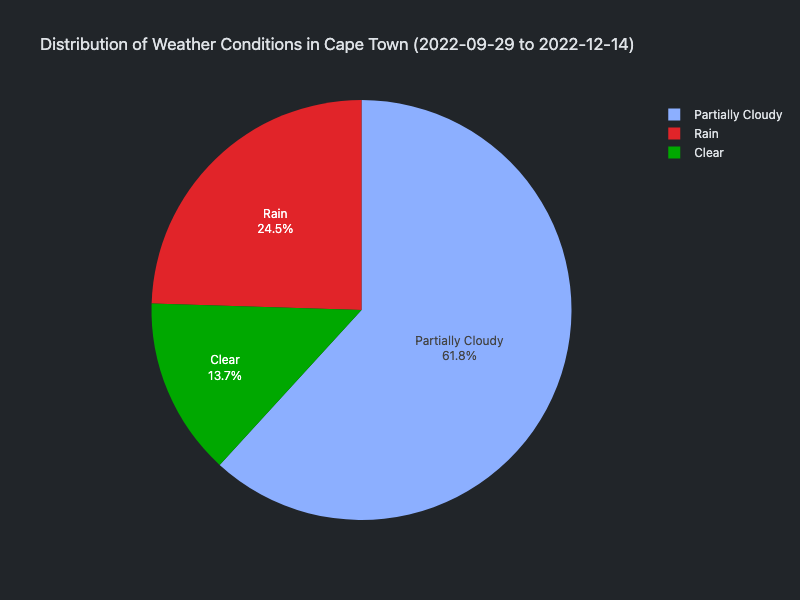}
        \small Chart B (Right)
    \end{minipage}

    \vspace{1em}

    \begin{minipage}[t]{0.98\textwidth}
        \small
        \textbf{Comparison Summary:} A comparison of Cape Town's weather distributions reveals a clear transition toward drier and brighter conditions between the June-August 2022 and September-December 2022 periods. The most notable shift is the significant decline in rainy days, which dropped from 37.7\% in the earlier period to 24.5\% in the later months. Concurrently, partially cloudy conditions surged to become even more dominant, increasing from a 48.2\% share to 61.8\%. Clear days also experienced a slight uptick, rising from 11.4\% to 13.7\%. Additionally, while overcast conditions made up 2.7\% of the weather in the first period, they disappeared entirely from the distribution during the latter timeframe.
    \end{minipage}
    \caption{
    An example pair of pie charts.
    }\label{appendix_fig:chartdiff_example_pie}
\end{figure*}

\begin{figure*}[t]
    \centering

    \begin{minipage}[t]{0.49\textwidth}
        \centering
        \includegraphics[width=\linewidth]{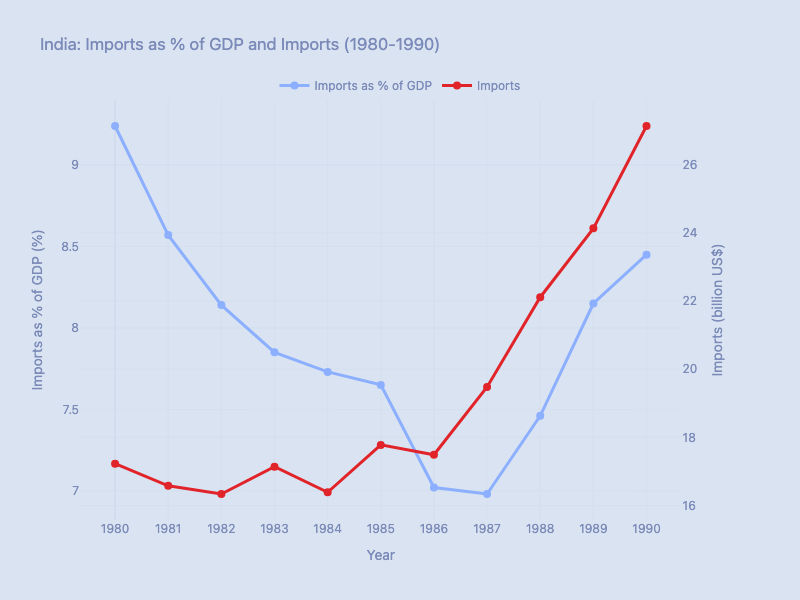}
        \small Chart A (Left)
    \end{minipage}
    \hfill
    \begin{minipage}[t]{0.49\textwidth}
        \centering
        \includegraphics[width=\linewidth]{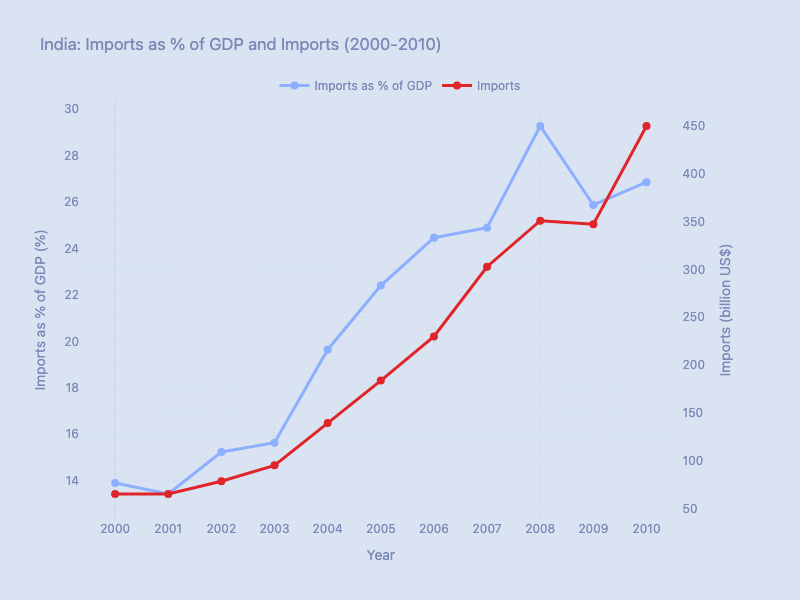}
        \small Chart B (Right)
    \end{minipage}

    \vspace{1em}

    \begin{minipage}[t]{0.98\textwidth}
        \small
        \textbf{Comparison Summary:} Between 1980 and 1990, India's imports were relatively stagnant, with absolute values growing modestly from \$17.23 billion to \$27.13 billion and the import share of GDP dipping mid-decade before recovering to 8.45\%. In stark contrast, the 2000 to 2010 period exhibited explosive growth, with absolute imports surging from \$65.12 billion to nearly \$450 billion. Similarly, India's imports as a percentage of GDP nearly doubled during the 2000s, climbing rapidly from 13.9\% in 2000 to a peak of 29.27\% in 2008. While the 1980s data shows mostly flat trends with only mild late-decade growth, the 2000s dataset reveals a powerful upward trajectory that was only briefly interrupted by a minor contraction in both metrics in 2009. Ultimately, the comparison illustrates a massive shift in economic scale, highlighting India's drastically increased integration into global trade during the 2000s compared to its low import reliance in the 1980s.
    \end{minipage}
    \caption{
    An example pair of multi-series line charts.
    }\label{appendix_fig:chartdiff_example_linem_multiscale}
\end{figure*}

\begin{figure*}[t]
    \centering

    \begin{minipage}[t]{0.49\textwidth}
        \centering
        \includegraphics[width=\linewidth]{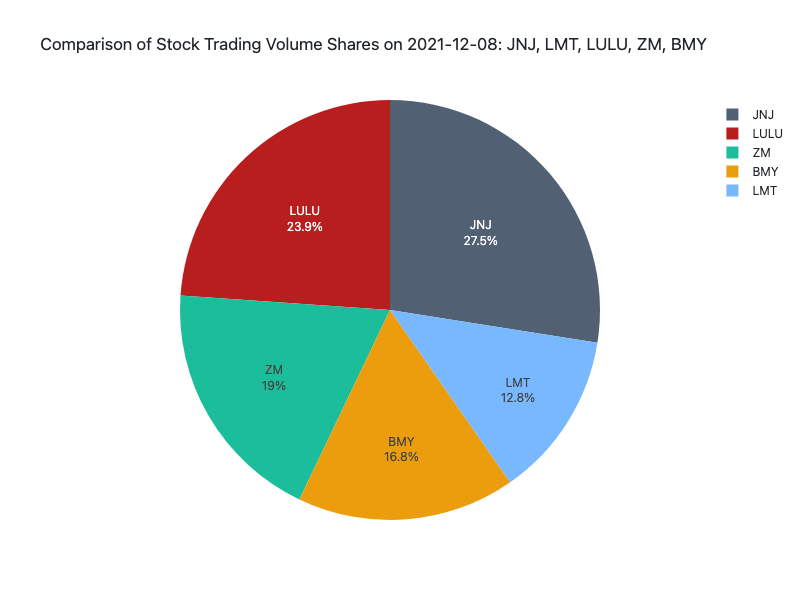}
        \small Chart A (Left)
    \end{minipage}
    \hfill
    \begin{minipage}[t]{0.49\textwidth}
        \centering
        \includegraphics[width=\linewidth]{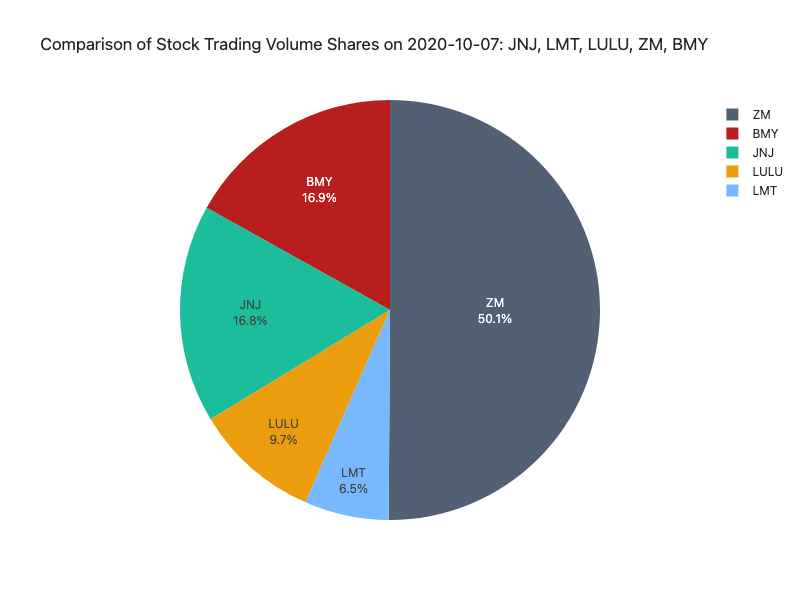}
        \small Chart B (Right)
    \end{minipage}

    \vspace{1em}

    \begin{minipage}[t]{0.98\textwidth}
        \small
        \textbf{Comparison Summary:} A comparison of stock trading volume shares between October 7, 2020, and December 8, 2021, reveals a dramatic shift in market focus among the five observed entities. Most notably, ZM dominated the trading volume in 2020 with a massive 50.1\% share but experienced a steep decline to just 19.0\% by late 2021. Conversely, JNJ and LULU absorbed much of this redistributed volume, with JNJ's share surging from 16.8\% to a leading 27.5\%, and LULU's more than doubling from 9.7\% to 23.9\%. LMT also saw a notable increase, nearly doubling its share from 6.5\% to 12.8\% over the same period. Meanwhile, BMY was the only remarkably stable stock in the group, maintaining a consistent share of roughly 16.8\% across both dates.
    \end{minipage}
    \caption{
    An example pair of pie charts.
    }\label{appendix_fig:chartdiff_example_linem_multiscale}
\end{figure*}

\clearpage
\section{Prompt templates}
\label{sec:instructions_prompts}

\begin{figure*}[h]   
\centering
\begin{tcolorbox}[
    enhanced,
    colback=white,        
    colframe=blue!60!black,   
    colbacktitle=blue!60!black, 
    coltitle=white,       
    fonttitle=\bfseries\large,  
    title= Prompt Template for Generating Candidate Annotations,
    attach boxed title to top center={
        yshift=-2mm
    },
    boxed title style={
        sharp corners,
        boxrule=0pt,
    },
    top=6mm,
    arc=2mm,              
    boxrule=0.8pt,
]
You are a professional data analyst.

You will compare two small datasets (CSV format) that describe one of the following:

\qquad1. Data of the same entity across two time ranges

\qquad2. Data of two entities across the same time range

\qquad3. Data of two entities across two time ranges

\qquad4. Multiseries data of the same entity across two time ranges

\qquad5. Multiseries data of two entities across the same time range

\qquad6. Comparison of multiple entities' shares across two time ranges

\qquad7. Comparison of two entities' shares across the same time range

\qquad8. Comparison of the same entity's shares across two time ranges

\qquad

Your task is to identify the main differences between the datasets in terms of trends, fluctuations, or anomalies.

Your response should be concise, accurate, and informative.

\qquad

Dataset A:

\qquad\texttt{$<$CSV\_A$>$}

Dataset B:

\qquad\texttt{$<$CSV\_B$>$}

\qquad

Write your comparison as a single cohesive paragraph of no more than five sentences.
Do not use bullet points or lists.

\end{tcolorbox}
\caption{Prompt template for generating candidate annotations.}
\label{prompt_for_candidate_annotations}
\end{figure*}

\begin{figure*}[t]   
\centering
\begin{tcolorbox}[
    enhanced,
    colback=white,        
    colframe=blue!60!black,   
    colbacktitle=blue!60!black, 
    coltitle=white,       
    fonttitle=\bfseries\large,  
    title= Prompt Template for Judging Candidate Annotations,
    attach boxed title to top center={
        yshift=-2mm
    },
    boxed title style={
        sharp corners,
        boxrule=0pt,
    },
    top=6mm,
    arc=2mm,              
    boxrule=0.8pt,
]
You are an expert evaluator for chart-comparison annotations.

You will receive:

\qquad1. Dataset A in CSV format

\qquad2. Dataset B in CSV format

\qquad3. A candidate comparison summary

\qquad

Your task is to decide whether the candidate summary should be accepted as a valid annotation.

Judge the summary ONLY against the CSV data.

\qquad

Accept the summary only if:

\qquad- it is factually supported by the data

\qquad- it captures the main differences between the datasets

\qquad- it does not omit the dominant trend, anomaly, ranking change, or share change

\qquad- it is clear and specific rather than generic

\qquad

Reject the summary if:

\qquad- it contains any material factual error

\qquad- it confuses Dataset A and Dataset B

\qquad- it invents unsupported claims

\qquad- it is too vague to be useful

\qquad- it misses an important difference shown in the data

\qquad

Material factual errors include:

\qquad- wrong trend direction

\qquad- wrong relative comparison

\qquad- wrong anomaly/peak/trough/crossover

\qquad- wrong share or ranking statement

\qquad- incorrect entity/category/time interpretation

\qquad

Return only one word:

\qquad ACCEPT

or

\qquad REJECT

\qquad

Dataset A:

\qquad\texttt{$<$CSV\_A$>$}

Dataset B:

\qquad\texttt{$<$CSV\_B$>$}

Candidate summary:

\qquad\texttt{$<$CANDIDATE\_SUMMARY$>$}

\end{tcolorbox}
\caption{Prompt template for judging candidate annotations.}
\label{prompt_for_review_annotations}
\end{figure*}

\begin{figure*}[t]   
\centering
\begin{tcolorbox}[
    enhanced,
    colback=white,        
    colframe=blue!60!black,   
    colbacktitle=blue!60!black, 
    coltitle=white,       
    fonttitle=\bfseries\large,  
    title= Prompt Template for Generating Comparison Summaries,
    attach boxed title to top center={
        yshift=-2mm
    },
    boxed title style={
        sharp corners,
        boxrule=0pt,
    },
    top=6mm,
    arc=2mm,              
    boxrule=0.8pt,
]
You are a professional data analyst.

\qquad

Compare Chart A (left) and Chart B (right) using only the information visible in the charts.

\qquad

Focusing on how their data differs in terms of overall trends, fluctuations, and any notable anomalies and emphasize the most important contrasts between the two charts rather than describing each chart independently.

\qquad

Your analysis must be concise, accurate, and written as a single cohesive paragraph of no more than five sentences.

\qquad

Avoid bullet points, lists, or redundant phrasing.

\qquad

\texttt{$<$PAIR\_IMAGE$>$}
\end{tcolorbox}
\caption{Prompt template for generating comparison summaries.}
\label{prompt_for_genrate_predictions}
\end{figure*}

\begin{figure*}[t]   
\centering
\begin{tcolorbox}[
    enhanced,
    colback=white,        
    colframe=blue!60!black,   
    colbacktitle=blue!60!black, 
    coltitle=white,       
    fonttitle=\bfseries\large,  
    title= Prompt Template for Generating Comparison Summaries in Pipeline Methods,
    attach boxed title to top center={
        yshift=-2mm
    },
    boxed title style={
        sharp corners,
        boxrule=0pt,
    },
    top=6mm,
    arc=2mm,              
    boxrule=0.8pt,
]
You are a professional data analyst.

\qquad

Compare table A and table B using only the information in the tables.

\qquad

Focusing on how their data differs in terms of overall trends, fluctuations, and any notable anomalies and emphasize the most important contrasts between the two tables rather than describing each table independently.

\qquad

Your analysis must be concise, accurate, and written as a single cohesive paragraph of no more than five sentences.

\qquad

Avoid bullet points, lists, or redundant phrasing.

\qquad

Table A:

\qquad\texttt{$<$TABLE\_A$>$}

Table B:

\qquad\texttt{$<$TABLE\_B$>$}
\end{tcolorbox}
\caption{Prompt template for generating comparison summaries in pipeline methods.}
\label{prompt_for_genrate_predictions_in_pipeline}
\end{figure*}

\begin{figure*}[t]   
\centering
\begin{tcolorbox}[
    enhanced,
    colback=white,        
    colframe=blue!60!black,   
    colbacktitle=blue!60!black, 
    coltitle=white,       
    fonttitle=\bfseries\large,  
    title=Prompt Template for Generating LLM Random Guesses,
    attach boxed title to top center={
        yshift=-2mm
    },
    boxed title style={
        sharp corners,
        boxrule=0pt,
    },
    top=6mm,
    arc=2mm,              
    boxrule=0.8pt,
]
You are a professional data analyst.

You will compare two small datasets (CSV format) that describe one of the following:

\qquad1. Data of the same entity across two time ranges

\qquad2. Data of two entities across the same time range

\qquad3. Data of two entities across two time ranges

\qquad4. Multiseries data of the same entity across two time ranges

\qquad5. Multiseries data of two entities across the same time range

\qquad6. Comparison of multiple entities' shares across two time ranges

\qquad7. Comparison of two entities' shares across the same time range

\qquad8. Comparison of the same entity's shares across two time ranges

\qquad

Your task is to identify the main differences between the datasets in terms of trends, fluctuations, or anomalies.

Your response should be concise, accurate, and informative.

\qquad

Randomly guess a reasonable comparison based on the above instruction only as a single cohesive paragraph of no more than five sentences.

Directly write the comparison as if you access two small datasets.

Do not use bullet points or lists.                                                      

\end{tcolorbox}
\caption{Prompt template for generating random guesses from an LLM.}
\label{prompt_random_output}
\end{figure*}

\begin{figure*}[t]   
\centering
\begin{tcolorbox}[
    enhanced,
    colback=white,        
    colframe=blue!60!black,   
    colbacktitle=blue!60!black, 
    coltitle=white,       
    fonttitle=\bfseries\large,  
    title= Prompt Template for Generating GPT Score - Part 1,
    attach boxed title to top center={
        yshift=-2mm
    },
    boxed title style={
        sharp corners,
        boxrule=0pt,
    },
    top=6mm,
    arc=2mm,              
    boxrule=0.8pt,
]
You are an expert data analyst and evaluator.

You will receive:

\qquad1. Dataset A (CSV format), corresponding to Chart A (the left chart)

\qquad2. Dataset B (CSV format), corresponding to Chart B (the right chart)

\qquad3. A reference analysis (intended correct comparison)

\qquad4. A candidate analysis (to be evaluated)

Both analyses describe the differences between two charts derived from the datasets.

\qquad

Your task is to evaluate the quality of the candidate analysis.

\qquad

IMPORTANT PRINCIPLES:

\qquad- The datasets are the ultimate source of truth.

\qquad- The reference analysis is a guideline for expected coverage and importance, but it may contain minor imperfections.

\qquad- Do NOT reward surface similarity to the reference if the content is incorrect.

\qquad- Do NOT penalize the candidate for wording differences if the meaning is correct.

\qquad

Evaluation Procedure (follow internally, do not output):

\qquad1. First, analyze Dataset A and Dataset B to identify the key differences:

\qquad\qquad- overall trends (increasing, decreasing, stable)

\qquad\qquad- fluctuations (volatility, variability)

\qquad\qquad- notable anomalies (peaks, drops, outliers)

\qquad\qquad- major contrasts between the two datasets

\qquad

\qquad2. Check whether the reference analysis correctly reflects these differences.

\qquad\qquad- If the reference is partially incorrect, rely on the data instead.

\qquad

\qquad3. Evaluate the candidate analysis based on:

\qquad\qquad(a) Accuracy

\qquad\qquad\qquad- Are the statements factually consistent with the datasets?

\qquad\qquad\qquad- Any contradictions or incorrect claims should be heavily penalized.

\qquad\qquad(b) Completeness

\qquad\qquad\qquad- Does the candidate cover the main differences identified from the data?

\qquad\qquad\qquad- Missing minor details is acceptable, but missing key trends is not.

\qquad\qquad(c) Faithfulness

\qquad\qquad\qquad- Does the candidate avoid hallucinating patterns not supported by the data?

\qquad\qquad(d) Clarity

\qquad\qquad\qquad- Is the analysis coherent, concise, and easy to understand? 

(Continued in Figure~\ref{prompt_gpt_score_part_2})
\end{tcolorbox}
\caption{Prompt template (Part 1) for generating GPT Score.}
\label{prompt_gpt_score_part_1}
\end{figure*}

\begin{figure*}[t]   
\centering
\begin{tcolorbox}[
    enhanced,
    colback=white,        
    colframe=blue!60!black,   
    colbacktitle=blue!60!black, 
    coltitle=white,       
    fonttitle=\bfseries\large,  
    title= Prompt Template for Generating GPT Score - Part 2,
    attach boxed title to top center={
        yshift=-2mm
    },
    boxed title style={
        sharp corners,
        boxrule=0pt,
    },
    top=6mm,
    arc=2mm,              
    boxrule=0.8pt,
]
(Continued from Figure~\ref{prompt_gpt_score_part_1})

Scoring:

\qquad{G}ive a single integer score from 0 to 5:

\qquad\qquad5 = Excellent: Factually correct, captures all key differences, clear and concise

\qquad\qquad4 = Good: Mostly correct, minor omissions or small inaccuracies

\qquad\qquad3 = Fair: Partially correct, noticeable gaps or some incorrect statements

\qquad\qquad2 = Poor: Major errors or missing important trends

\qquad\qquad1 = Very poor: Mostly incorrect or largely irrelevant

\qquad\qquad0 = Fail: Completely incorrect, nonsensical, or empty

\qquad

\qquad{S}coring Rules:

\qquad\qquad- Prioritize factual accuracy over similarity to the reference.

\qquad\qquad- If the candidate contradicts the data, score must be $\leq$ 2.

\qquad\qquad- If the candidate misses the main trend, score must be $\leq$ 3.

\qquad\qquad- Minor wording or structure issues should NOT significantly reduce the score.

\qquad

Output Format:

\qquad{R}eturn ONLY a single integer (0-5).

\qquad{D}o NOT provide any explanation or additional text.

\qquad

{D}ataset A (Chart A - LEFT):

\qquad\texttt{$<$CSV\_A$>$}

{D}ataset B (Chart B - RIGHT):

\qquad\texttt{$<$CSV\_B$>$}

Reference analysis:

\qquad\texttt{$<$GROUND\_TRUTH\_ANNOTATION$>$}

Candidate analysis:

\qquad\texttt{$<$CANDIDATE\_ANALYSIS$>$} 
        
\end{tcolorbox}
\caption{Prompt template (Part 2) for generating GPT Score.}
\label{prompt_gpt_score_part_2}
\end{figure*}

\end{document}